# Parallax Effect Free Mosaicing of Underwater Video Sequence Based on Texture Features

Nagaraja S., Prabhakar C.J. and Praveen Kumar P.U.

Department of P.G. Studies and Research in Computer Science
Kuvempu University, Karnataka, India

## ABSTRACT

*In this paper, we present feature-based technique for construction of mosaic image from underwater video sequence, which suffers from parallax distortion due to propagation properties of light in the underwater environment. The most of the available mosaic tools and underwater image mosaicing techniques yields final result with some artifacts such as blurring, ghosting and seam due to presence of parallax in the input images. The removal of parallax from input images may not reduce its effects instead it must be corrected in successive steps of mosaicing. Thus, our approach minimizes the parallax effects by adopting an efficient local alignment technique after global registration. We extract texture features using Centre Symmetric Local Binary Pattern (CS-LBP) descriptor in order to find feature correspondences, which are used further for estimation of homography through RANSAC. In order to increase the accuracy of global registration, we perform preprocessing such as colour alignment between two selected frames based on colour distribution adjustment. Because of existence of 100% overlap in consecutive frames of underwater video, we select frames with minimum overlap based on mutual offset in order to reduce the computation cost during mosaicing. Our approach minimizes the parallax effects considerably in final mosaic constructed using our own underwater video sequences.*

## 1. INTRODUCTION

Image mosaicing is the process of assembling sequence of images of the same scene acquired at different angles and different timings. Mosaicing is a popular method to increase the field of view of a camera, by allowing many views of a same scene of different images combined into a single view. The underwater mosaic can be used in underwater environment applications such as geological and archaeological surveys, biology and environmental monitoring. Building mosaics of underwater images is a complex task that faces some specific challenges not present in aerial or terrestrial panorama generation due to non-uniform illumination in images and variations in color between images of a same scene. The lack of natural light in deep water and scattering phenomenon due to suspended particles affect the quality of images. The mosaic techniques can be classified into two categories. The first one is direct method [1][2][3][4] and second one is the feature-based method [5][6][7][8]. The direct method uses all the available video image data and heavily dependent on brightness constancy and initialization. Whereas feature -based method uses features of an image such as corners, junctions and blobs. The successful and popular feature-based mosaicing techniques cannot be applied directly on underwater images because of inherent nature of images and many problems posed by underwater environment.

The literature survey reveals that most of the researchers have employed feature-based techniques for constructing the underwater mosaic. The final mosaic of these early techniques shows some artifacts like ghosting, blurring, and seam, which are appeared due to parallax effect [9] which is common in underwater input images. The parallax effect is the difference in the apparent position of an object viewed along two different lines of sight measured by an angle of inclination [10]. The most of the early underwater mosaic techniques not considered parallax effect in input images and they assumed that scene is planar in order to calculate motion between images based on homography estimation. This is very rare case in underwater environment. Q Zhong Li et al. [11] have proposed technique for stitching pair of underwater images captured using multiple cameras. They demonstrate that underwater images are having parallax distortion and proposed technique minimize it. Initially, they determine significant parameters in the projective transformation matrix related to parallax distortion and adjust these parameters in order to minimize the parallax distortion. The final mosaic is still having parallax effects such as ghosting and small amount of seam. Another drawback is that while calculating projective transformation matrix between pair of images, they select corresponding feature points manually, which is very difficult task in real time underwater applications.  The parallax effect impacts both the registration and blending steps [10]. If two images suffering from parallax are successfully registered, there may be chance of misalignments appear, which is common in underwater images [10]. This problem can be solved by using image blending technique. The artifacts such as ghosting, blurring and seam arises due to misalignments during global image registration can be eliminated using local registration [1] [9].

In this paper, we proposed feature-based technique to construct parallax effect free image mosaic from underwater video sequence captured using monocular video camera. After global registration based on homography estimated using RANSAC, we perform local registration using multi-band blending function [12] in order to correct misalignments which yields parallax effect free mosaic. The success of feature-based image mosaicing is heavily depending upon the pairwise image registration based on establishing features correspondence. The colours of two underwater images can be considerably different from each other due to the propagation properties of light in underwater environment during imaging process. The variations in colour between two images of same scene affect the accuracy of registration which is important step in feature-based mosaicing. In our method, initially, we align colours between two selected images based on colour distribution adjustment [13]. The main challenge in mosaicing for underwater images is lack of features which is not comparable with out-of-water images.  In underwater images majority of the features are from concrete texture and sandy area and minority of the features belong to objects such as fishes or rocks. The most of the underwater mosaic methods adapted SIFT descriptor which uses features such as corners, junctions and blobs. It is very difficult to find these features in underwater images due to presence of concrete texture and sandy area. Therefore, SIFT based techniques does not yield robust and invariant features in order to find accurate features correspondence [14]. It is observed that the texture parameters that remain constant for the scene path for the whole underwater image sequence [15]. Thus, we extract the texture features using Centre Symmetric Local Binary Pattern (CS-LBP) descriptor from the keypoints detected using DoG [16]. In underwater, due to propagation properties of light in the underwater, the image acquisition rate is high compared to camera speed which leads to 100% overlap on consecutive frames. Because of overlap, it is clear that not all frames are needed to construct a mosaic that covers the whole scene. Therefore many similar frames can be discarded that have 10% to 20% overlap in order to reduce the computation cost. We select the frames based on their mutual offset is as large as possible [17] and the selected frames are stitched together. Finally, as we mentioned, the misalignments are corrected by using multi-band blending function.

The rest of the paper is organized as follows: Section 2 describes related work. Section 3 describes our approach for parallax effect free underwater image mosaic construction. The experimental results are presented in Section 4. The conclusion is given in section 5.

## 2. RELATED WORK

There are only few papers published on underwater image mosaics. Most widely used technique is feature-based mosaicing. The techniques can be classified into three categories based on applications.

*Mosaic for navigation:* Underwater vehicles are useful for exploration and monitoring of the seabed. N. Gracias et al. [18] developed a method for building video mosaics of the sea bottom for visual navigation of underwater robots based on feature based registration and topology estimation. J. Guo et al. [19] proposed mosaicing technique for underwater vehicle navigation based upon a Maximum a Posteriori estimation technique, combines a least-mean-squared-error estimator and a Kalman Filter. N. Gracia et al. [20] presented technique for the construction of underwater mosaics for underwater vehicle navigation. The motion estimation is based on an initial matching of corresponding areas over pairs of images. The motion models under the projective geometry framework, allow for the creation of high quality mosaics where no assumptions are made about the camera motion. Then they determine 3D position and orientation of a vehicle from new views of a previously created mosaic.

*Mosaic for environment monitoring:* H. Bagheri et al. [21] developed feature based method using SIFT for construction of underwater image mosaicing for counting benthic species from sea floor images. The images are captured using digital still camera and transformation matrix is estimated using RANSAC and finally local alignment is carried using Multiband blending. B. Gintert et al. [22] proposed underwater landscape mosaics for coral reef mapping and monitoring. The constructed mosaic from images covering area of an interest are captured using robot with multi camera imaging system can achieve wide field of view (FOV) of surroundings.

*Mosaic for visual mapping:* Within mapping applications, the maps obtained from optical data are becoming essential in different study areas such as biological, geological and archaeological surveys. A. Elibol et al. [23] have proposed image mosaicing for large area visual mapping in underwater environment using multiple robots. The authors proposed a new global alignment method which works on the mosaic frame and does not require non-linear optimisation. They identified overlapping image pairs in trajectories carried out by the different robots during the topology estimation process. Prados et al. [10] presented technique for construction of mosaics of large area. Initially, they applied illumination compensation function and contrast compensation function in order to correct illumination and contrast respectively. After images are enhanced, they perform feature based image registration in order obtain registration parameters. Images are sorted according to their diagonal length, and classified into subsets of similar altitudes. After selecting the images with considerable overlapping, the global registration is performed and finally Graph-cut based blending is done. This method is computationally expensive because it involves lot of successive steps.

## 3. UNDERWATER IMAGE MOSAICING

We proposed feature-based mosaicing technique to construct parallax effect free image mosaic from underwater video sequence captured using monocular video camera. After global registration based on homography estimated using RANSAC, we perform local registration using multi-band blending function in order to correct misalignments which yields parallax effect free mosaic.

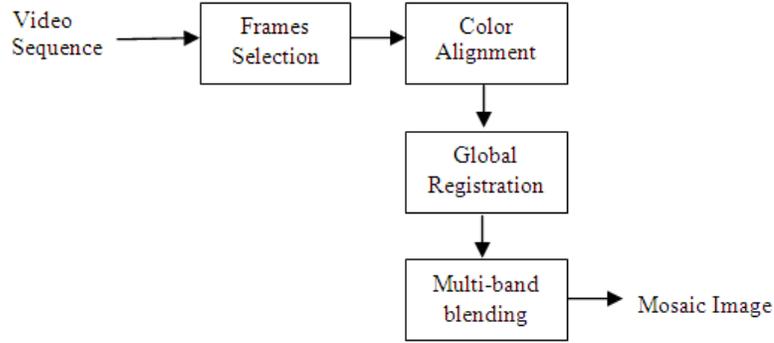

Figure 1: Flow diagram of our method.

## 3.1 Frames Selection

In underwater imaging, the image acquisition rate is high compared to camera speed which leads to 100% overlap on consecutive frames. Hence, it is clear that not all frames are needed to construct a mosaic that covers the whole scene. Therefore many similar frames can be discarded in order to reduce the computation cost. For instance, the fixed scene is captured while moving the video camera with only horizontal forward translation (no vertical movement), the ROI (Image patch) projected opposite to camera motion i.e. towards left position in the consecutive frames. This phenomenon is shown the Figure 2. Consider set of n consecutive frames, here it is assumed that the first frame is reference and remaining n-1 frames are target frames. It is observed that mutual offset between reference frame and its successive frame (target frame 1) is very less compared mutual offset for target frame 3. Similarly mutual offset for target frame n is very high compared to other target frames. Hence, we can conclude that less mutual offset between two frames means that proportion of overlapping is very high; on the other hand, higher the mutual offset means lower the overlapping. Hence, we select the frames based on mutual offset is as large as possible [17] and the selected frames are stitched together. We employed block based stereo correspondence technique to measure the mutual offset between pair of frames. The block based stereo correspondence method for calculating the disparity map is to use small region of pixels in the right frame (reference), and searching for the closest matching region of pixels in the left frame (target).

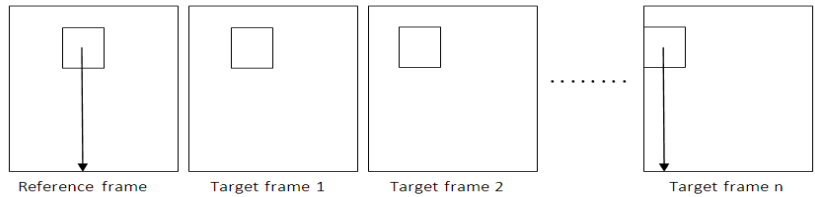

Figure 2: Demonstration of finding mutual offset between frames

While searching in the target frame, we start at the same coordinates as our image patch and search to the left and right up to some maximum distance (disparity range). The similarity metric for finding the closest matching block by employing simple operation called sum of absolute differences (SAD) is defined as

$$SAD = \sum_{i=1}^{n} \sum_{j=1}^{n} |t_{i,j} - s_{i,j}|, \qquad (1)$$

where $s_{i,j}$ are the elements of reference block, $t_{i,j}$ are the elements of the target block and *n* size of the block. In order to compute the sum of absolute differences between the blocks of reference and target frame, we subtract each pixel in the block of reference frame from the corresponding pixel in the block of target frame and take the absolute value of the differences. Finally we sum up all of these differences, which yield single value that measures the similarity between the two image patches. Lower the value means the patches are more similar. After finding the corresponding patch in the target frame, we compute the disparity between reference image patch and target image patch. We select the frame which is having less overlapping compared to other target frames based on highest disparity value. Finally, the frames are preferred that are farther away from the previous selected frame. This process is repeated for whole set of frames, which yields subset of frames with less overlapping between consecutive frames.

## 3.2 Colour Alignment

In the frame selection stage, we selected the subset of frames with less overlapping and the consecutive pair of frames can be considerably different from each other due to the propagation properties of light in underwater environment during imaging process. The variations in colour between two frames of same scene affect the accuracy of registration which is important step in feature-based mosaicing. In this method, initially, we align colours between two consecutive frames in subset of frames based on colour distribution adjustment using simple statistical analysis to implement colour alignment of the frames. First compute the mean and standard deviation of the intensity in each RGB channel for each frame. Then, we align the colour of one frame to other frame using these values. To align the colours of an source frame to match with another frame i.e. target frame, the following colour mapping equation for each RGB channel is defined as

$$I_{C,s,A} = \frac{\sigma_{C,t}}{\sigma_{C,s}}\left(I_{C,s} - m_{C,s}\right) + m_{C,t} \ , \qquad (2)$$

where *C* is each colour channel, $I_C$ is the image intensity of each pixel, $m_C$ is the mean, $\sigma_C$ is the standard deviation, A is aligned frame and s and t are source and target respectively. The above colour alignment makes the two underwater consecutive frames have the same mean and standard deviation of intensity for each channel.

## 3.3 Image Registration

Image registration is the process of establishing mapping between two or more images and aligning them with respect to a common co-ordinate system. We used feature-based image registration in order to align the consecutive pair of frames. We detect robust and discriminative feature points using Difference of Gaussian (DoG) technique, which is part of SIFT and widely used in various applications. For each feature or interest point, a texture descriptor is built using CS-LBP [25] technique to distinctively describe the local region around the interest point. Then feature descriptors are matched using Nearest Neighbour Distance Ratio (NNDR) to measure the similarity. Homography is estimated using random sample consensus (RANSAC) from the pairs of matched locations.

*Detection of Interest points using DoG*: The scale invariant features are detected from a reference image using Difference of Gaussian (DoG) technique. These features are invariant to image translation, rotation, scaling, illumination, viewpoint, noise etc.

*CS-LBP Descriptor*: Texture classification is one of the important problems in texture analysis; it has been studied for several decades and shows significant improvements. Ojala et al.[24]

proposed a popular texture descriptor called local binary pattern (LBP), which is good texture discriminative property. LBP has a property that favours its usage in interest region description such as tolerance against illumination changes, computationally simple and efficient. The main drawbacks of LBP are, it produces long histogram, this is difficult to use in the context of a region descriptor and is not too robust on flat areas of the images. In order to overcome these drawbacks M. Heikkila et al. [25] introduced Centre Symmetric Local Binary Pattern (CS-LBP) is extended version of LBP. CS-LBP is robust on flat image areas, generally found in underwater environment. The CS-LBP descriptor outperforms the existing local descriptor for most of the test cases, especially for images with severe illumination variations and it captures better gradient information than original LBP.

CS-LBP compares centre symmetric pairs of pixels in a local region instead of computing of local binary pattern in a local region, each neighbour pixel compare with centre pixel of local region. It computes halves the number of comparisons for the same number of the neighbourhoods of the local region. CS-LBP produce only 16 different binary patterns compared to 256 different binary patters of LBP for 8 neighbourhoods. It is a robust on flat regions obtained by thresholding the gray level differences with a small value t and defined mathematically as

$$CS - LBP_{r,p,t} = \sum_{i=0}^{(p/2)-1} S(g_i - g_{i+(p/2)})2^i, \tag{3}$$

$$S(x) = \begin{cases} 1 & x > t \\ 0 & x \le t \end{cases}, \tag{4}$$

where $g_i$ and $g_{i+(p/2)}$ are the gray values of the centre symmetric pairs of pixels of $p$ equally spaced pixels in a circle of radius r. It also noticed that CS-LBP is closely related to gradient operator, because it considered gray value differences between pairs of opposite pixels in neighbourhood.

In order to incorporate spatial information into the descriptor, the input region is divided into cells with Cartesian grid 4 x 4 (16 cells) is used. For each cell a CS-LBP histogram is built and the resulting descriptor is a 3D histogram of CS-LBP feature values and locations. The number of different feature values ($2^{p/2}$) depends on the neighbourhood size $p$ of the chosen CS-LBP. The final descriptor is formed $m$ x $p$ x $2^{p/2}$ dimensional vectors by concatenating the feature histograms computed for the cells, where $m$ is the gird size and $p$ is the CS-LBP neighbourhood size. We compute the similarity between the descriptors using Nearest Neighbour Distance Ratio (NNDR) based technique.

*Homography estimation*: The Random sample Consensus (RANSAC) algorithm proposed by Fischler and Bolles [26] is a general parameter estimation approach designed to handle with a large proportion of outliers in the input data. RANSAC is a resampling technique that generates candidate solutions by using the minimum number of data points required to estimate the underlying parameters of the model. The number of iterations $N$ is chosen high enough to ensure that the probability $p$ that at least one of the sets of random samples does not include an outlier. Let $u$ represent the probability that any selected data point is an inliers and $v=1-u$ the probability of observing an outlier. $N$ iterations of the minimum number of point's denoted m are required, where

$$1 - p = (1 - u^m)^N, \tag{5}$$

and thus with some manipulation

$$N = \frac{\log(1-p)}{\log(1-(1-v)^m)}.$$ (6)

### 3.4 Multi Band Blending

The computed homography matrix describes the 2D transformations between given pair of frames. We assumed that scene is planar because our aim is to construct mosaic of sea floor which is inherently flat in nature. Another assumption is that the camera moves in horizontal translation. Since underwater video sequence suffers from parallax, which impact the registration step. When registering a pair of frames suffering from parallax, the estimated homography may represent the dominant motion between both views. This leads to misalignments when overlying both views, resulting in artifacts like ghosting or broken image structures. Because of these problems a good blending technique is important. The Burt and Adelson [12] proposed technique called multi-band blending or pyramid blending, it is proved effective for image mosaicing without ghosting and blurring effects.

The plan of multi-band blending is to combine low frequencies over a large spatial range and high frequencies over a short range. The underwater image is decomposed into a number of N band pass images using the Laplacian pyramid. The Lapalacian pyramid of the final underwater image is formed using this equation.

$$L_k(i,j) = P_{1,k}(i,j)G_k(i,j) + P_{2,k}(i,j)(1 - G_k(i,j)),$$ (7)

where $P_{1,k}$ and $P_{2,k}$ are the $k^{th}$ level of Lapalacian pyramid decomposition for the two underwater images after coordinate adjustment, $L_k$ is the $k^{th}$ level of Lapalacian pyramid decomposition for the final combination of result. Similarly $G_k$ is the $k^{th}$ level of Gaussian pyramid decomposition of the image mask. Multi-band blending gradually blends the lower frequencies of the images while maintaining a sharper transition for the higher frequencies, which makes underwater image mosaicing is clearer.

## 4. EXPERIMENTAL RESULTS

We conducted experiments in order to evaluate our method on underwater video sequence which suffers from parallax distortion. The underwater video sequences are captured using water proof camera which is Canon-D10. We capture the video sequences at a distance 2 meter from surface of water. We kept some objects in water tank and small pond with turbid water and captured four sets of video sequences in different water conditions. The sample frames are shown in the Figure 2.

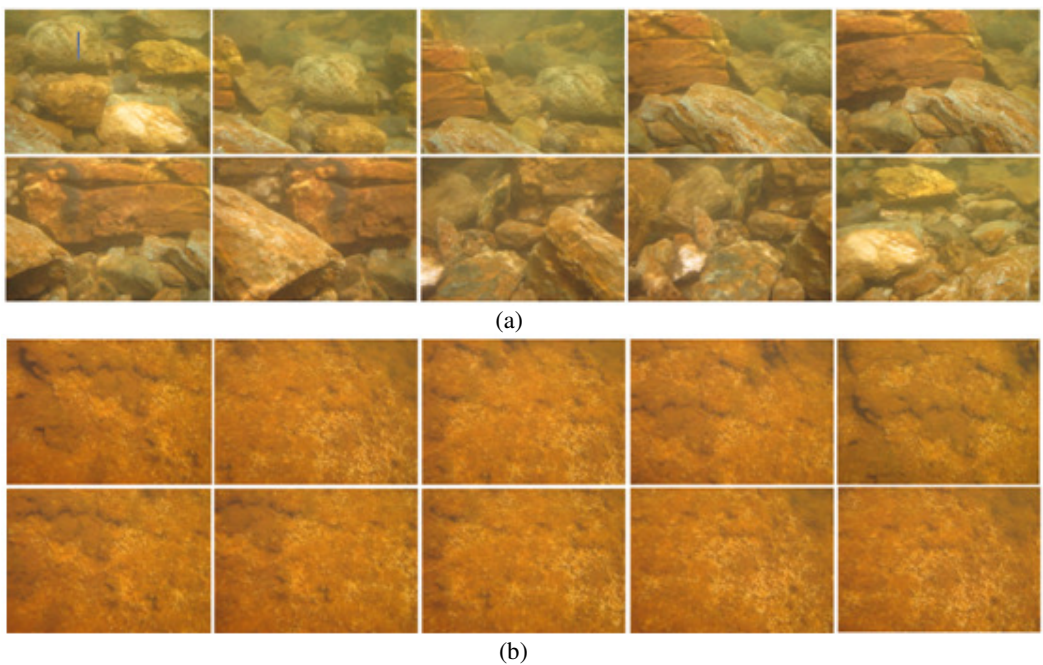

(a)

(b)

Figure 2: Some of the sequence of frames dataset (a) and (b) used for mosaic construction

The captured video frames are suffers parallax distortion, due to propagation properties of light in the underwater environment. From each video sequence, subset of frames was selected based on the frames selection criteria explained in the section 3.1, the subset of frames was further used for construction of image mosaic. We align colours between two selected frames using the simple statistical analysis method based on colour distribution adjustment i.e. explained in section 3.2.

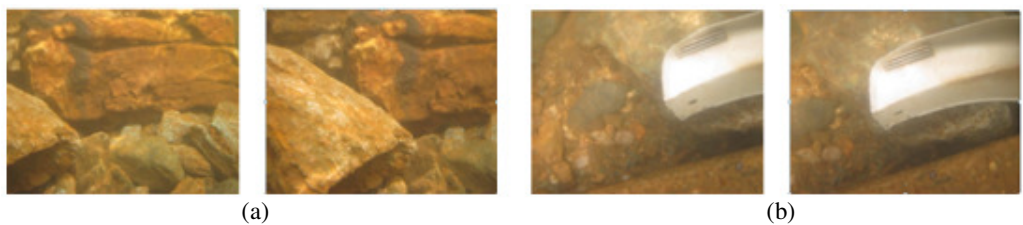

(a)                                        (b)

Figure 3: Color aligned pairs of frames (a) and (b) selected based on mutual offset

In order to find features correspondences, the first step is detection of reliable discriminate features using DoG technique and in order to extract the texture features from detected key points, we employed Centre Symmetric Local Binary Pattern (CS-LBP) descriptor. The matched points between pair of frames are used to estimate the homography using RANSAC. For each pair of frames, the homographies are estimated using at least 4 pairs of corresponding points. After global alignment of the frames, some pixel may not align properly this may cause blur or ghosting. We perform local registration using multi-band blending function in order to correct misalignments which yields parallax effect free mosaic.

## 4.1 Evaluation of Feature correspondences

We evaluated the accuracy of our method for detection of feature points using repeatability criteria. The comparison results of our method with existing methods are illustrated in the Table1.

From the results it is observed that, the repeatability rate of our method is very high compared to existing methods.This is due to the fact that we carried out pre-processing such as color alignment of frames, which aligns the frames with uniform color. This boost the performance of feature detector and another factor is that we used DoG method for feature detection.

Table 1: Comparison of repeatability measure of our approach with exiting feature detectors for the data set (a) has shown in the Figure 3.

| Approaches | Left Frame keypoints | Right Frame keypoints | Number of Matches | Repeatability |
|------------|---------------------|----------------------|-------------------|---------------|
| KLT[28] | 243 | 257 | 22 | 0.044 |
| Hessian[29] | 238 | 252 | 24 | 0.048 |
| Our approach | 262 | 274 | 31 | **0.058** |

We evaluate the performance of our approach for feature matching using Recall and Precision measures. The Table 2 shows the comparison of evaluation results with existing methods and it clearly demonstrates that our approach has high recall and precision compared to other methods. Since, texture information remains constant in the sequence of frames, the texture information used as description for detected key points enhances the feature matching accuracy. The texture information extracted using CS-LBP is robust to illumination variation and also contains gradient information.

Table 2: Comparison of Recall and Precision values for the data set (a) has shown in the Figure 3.

| Methods | Recall | 1-Precision |
|---------|--------|-------------|
| SIFT[16] | 0.62 | 0.58 |
| SURF[30] | 0.68 | 0.61 |
| Our Approach | **0.71** | **0.64** |

## 4.2 Comparison of mosaic results

We compared image mosaicing results of our approach with auto-mosaicing method [27]. Since there is no evaluation criterion available in the literature to evaluate the mosaic results, we compared the results visually. The Figure 4 and Figure 5 shows the mosaic results constructed using sequence of frames of datasets (a) and (b) shown in the Figure 2. Our approach yields good image mosaic without any artifacts, ghosting and blurring. However, auto-mosaic method failed to produce good mosaic and results suffer from small amount of ghosting and blurring as indicated by the red circle in the Figure 4(b) and 5(b). Auto-mosaicing method could not align underwater images properly using global transformation, therefore the mosaicing results suffer from seam, blurring and ghosting as shown in the Figure 4, 5, 6 and 7. Our feature-based mosaicing technique minimizes the parallax effects by adopting an efficient local alignment technique after global registration.

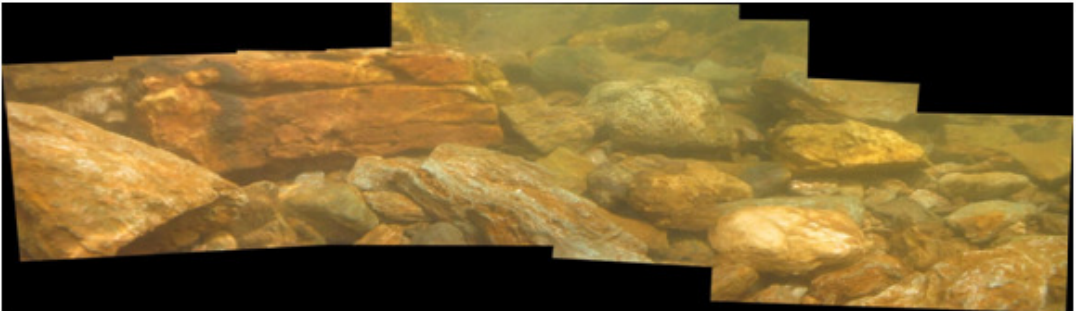

(a)

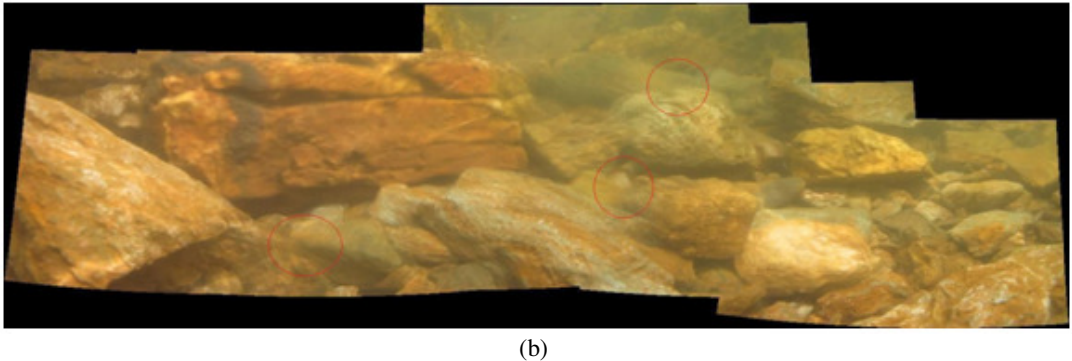

(b)

Figure 4: Comparison of mosaic results: (a) Our approach (b) Auto-mosaicing technique (red circle indicates blurring in final result)

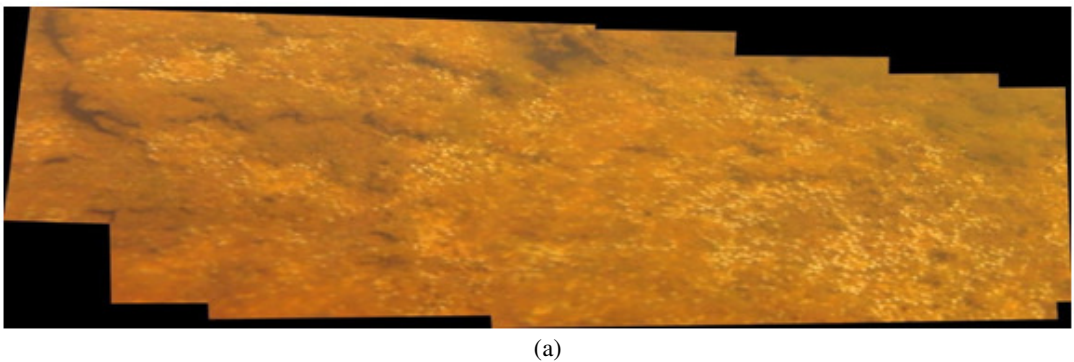

(a)

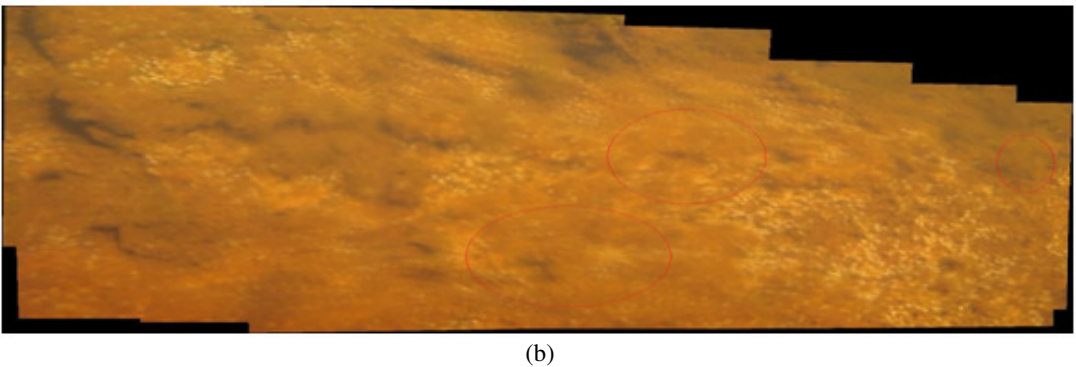

(b)

Figure 5: Comparison of mosaic results: (a) Our approach (b) Auto-mosaicing technique (red circle indicates blurring in final result)

## 5. CONCLUSIONS

In this paper, we proposed parallax effect free mosaic for underwater video sequence based on texture features. Since there is no benchmark database is available for underwater video sequence, we conducted experiments using our own captured video sequences. After global registration based on homography estimated using RANSAC, we perform local registration using multi-band blending function in order to correct misalignments which yields parallax effect free mosaic. The experimental results shows that, the combination DoG with CS-LBP descriptor yields high repeatability, recall and precision rate compared to other methods. We evaluate image mosaicing results of our approach with auto-mosaicing method results and our technique yields parallax effect free image mosaic with good quality. Our approach consumes less processing time compared to auto-mosaic method, this is because we selected subset of frames based on frames selection criteria and another fact is that usage of CS-LBP descriptor which involves simple arithmetic calculation and it produces only 16 bins compared to 256 bins of LBP. Based on these observations, we conclude that our technique construct good underwater image mosaic, which can be used for underwater navigation, visual mapping and other underwater applications.

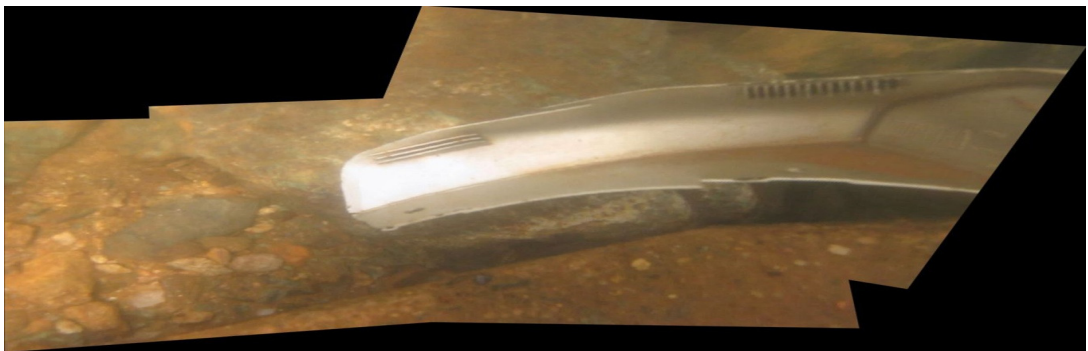

(a)

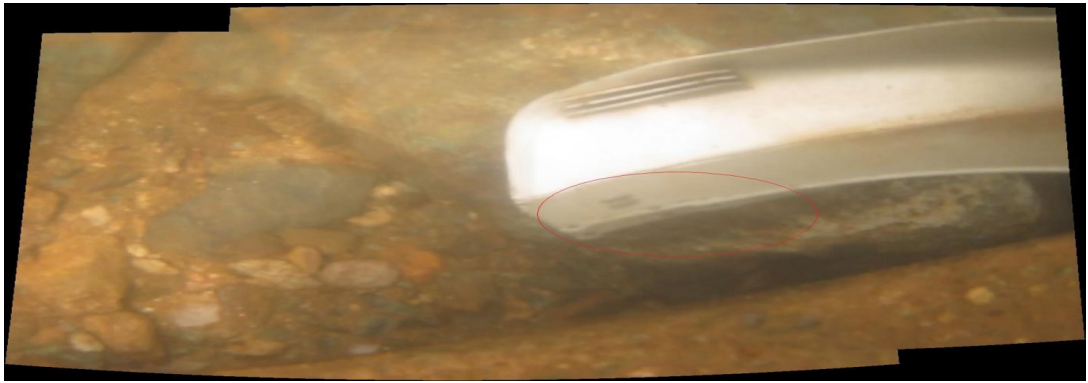

(b)

Figure 6: Comparison of mosaic results: (a) Our approach (b) Auto-mosaicing technique (red circle indicates ghosting in final result)

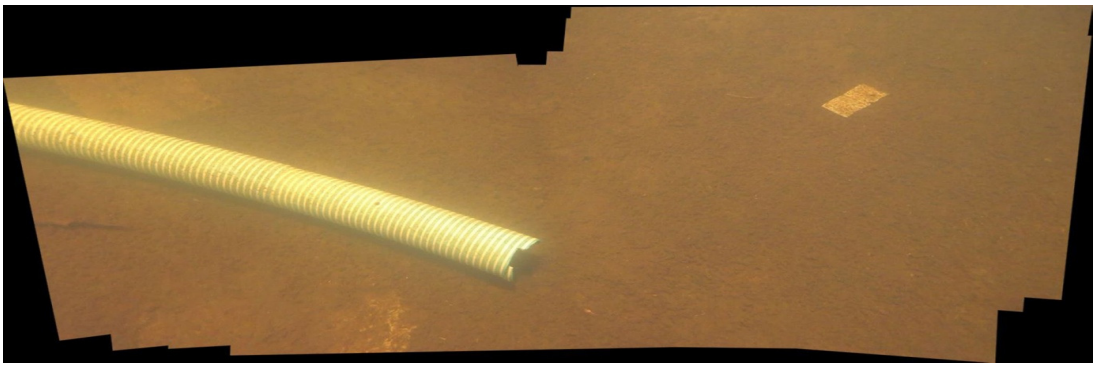

(a)

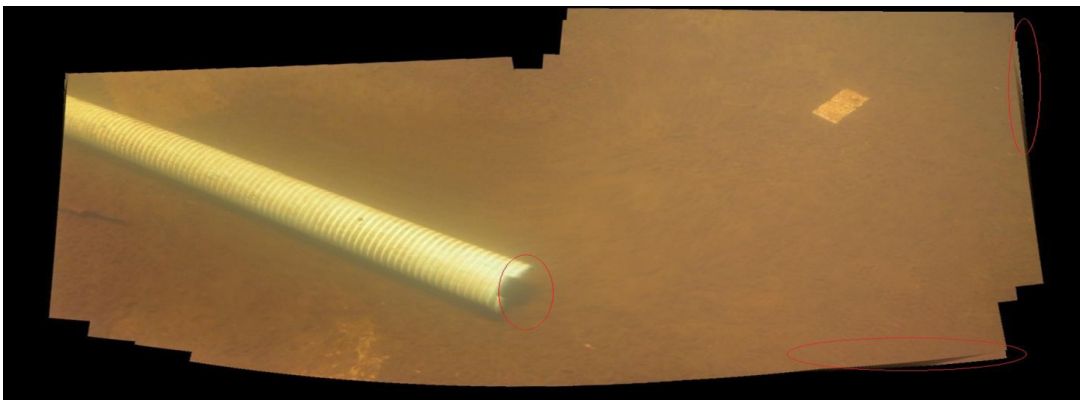

(b)

Figure 7: Comparison of mosaic results: (a) Our approach (b) Auto-mosaicing technique (red circle indicates seam and blur in final result)